\documentclass[runningheads]{llncs}
\usepackage{todonotes}
\usepackage{amsmath}
\usepackage{amssymb}
\usepackage{bbm}
\usepackage{adjustbox}
\usepackage{array,colortbl,multirow,multicol,booktabs,ctable}

\usepackage[title]{appendix}
\usepackage{xcolor}
\usepackage{setspace}
\usepackage{graphicx}
\graphicspath{{images/}}
\usepackage{threeparttable}
\usepackage[inline,shortlabels]{enumitem}  
\usepackage{hyperref}
\usepackage{url}
\makeatletter
\makeatletter
\def\Url@twoslashes{\mathchar`\/\@ifnextchar/{\kern-.2em}{}}
\g@addto@macro\UrlSpecials{\do\/{\Url@twoslashes}}
\makeatother

\newcommand{\secref}[1]{\S\ref{#1}} 
\newcommand{\appref}[1]{Appendix~\ref{#1}}

\newcommand{\figref}[1]{Fig.~\ref{#1}}    
\newcommand{\tabref}[1]{Table~\ref{#1}}  
\newcommand{\equref}[1]{eq.~(\ref{#1})}

\newcommand{\eg}{e.g., }

\usepackage{color}

\DeclareMathOperator{\MLP}{MLP}
\DeclareMathOperator{\BERT}{BERT}

\begin{document}

\title{JobBERT:\\Understanding Job Titles through Skills}
%
%
\author{Jens-Joris~Decorte\inst{1,2} \and
Jeroen~Van~Hautte\inst{2} \and
\\ Thomas~Demeester\inst{1}
\and
Chris~Develder\inst{1}
}
\authorrunning{J.-J.~Decorte \emph{et al.}}
%
\institute{Ghent University -- imec, 9052 Gent, Belgium \\
\urlstyle{rm}
\email{\{jensjoris.decorte, thomas.demeester, chris.develder\}@ugent.be}\\
\url{https://ugentt2k.github.io/}
\and
TechWolf, 9000 Gent, Belgium\\
\email{\{jensjoris, jeroen\}@techwolf.ai} \\
\url{https://techwolf.ai/}}
\maketitle              
\begin{abstract}

Job titles form a cornerstone of today's human resources (HR) processes. 
Within online recruitment, they allow candidates to understand the contents of a vacancy at a glance, while internal HR departments use them to organize and structure many of their processes. 
As job titles are a compact, convenient, and readily available data source, modeling them with high accuracy can greatly benefit many HR tech applications. 
In this paper, we propose a neural representation model for job titles, by augmenting a pre-trained language model with co-occurrence information from skill labels extracted from vacancies.
Our JobBERT method leads to considerable improvements compared to using generic sentence encoders, for the task of job title normalization, for which we release a new evaluation benchmark.

\end{abstract}
%
%
%

\section{Motivation and Related Work}

Job titles form a crucial piece of textual information within the job market.
For job seekers, they are a primary way of searching for relevant job postings. In HR departments of large organizations, they serve as an accessible data source for gathering insights or taking actions concerning their in-house talent. Although job titles generally adhere to a certain structure~\cite{van-hautte-etal-2020-leveraging}, they are often freely entered, unstructured texts. 
Nevertheless, standardized lists of job titles have been established, \eg in the European Skills, Competences, Qualifications and Occupations (ESCO) taxonomy~\cite{ESCO}. 
The task of relating free-form job titles to their most relevant standardized job title is called job title normalization and is a crucial step towards valuable information extraction (IE) from HR data. 

\begin{figure}[ht]
\centering
\includegraphics[width=0.75\textwidth]{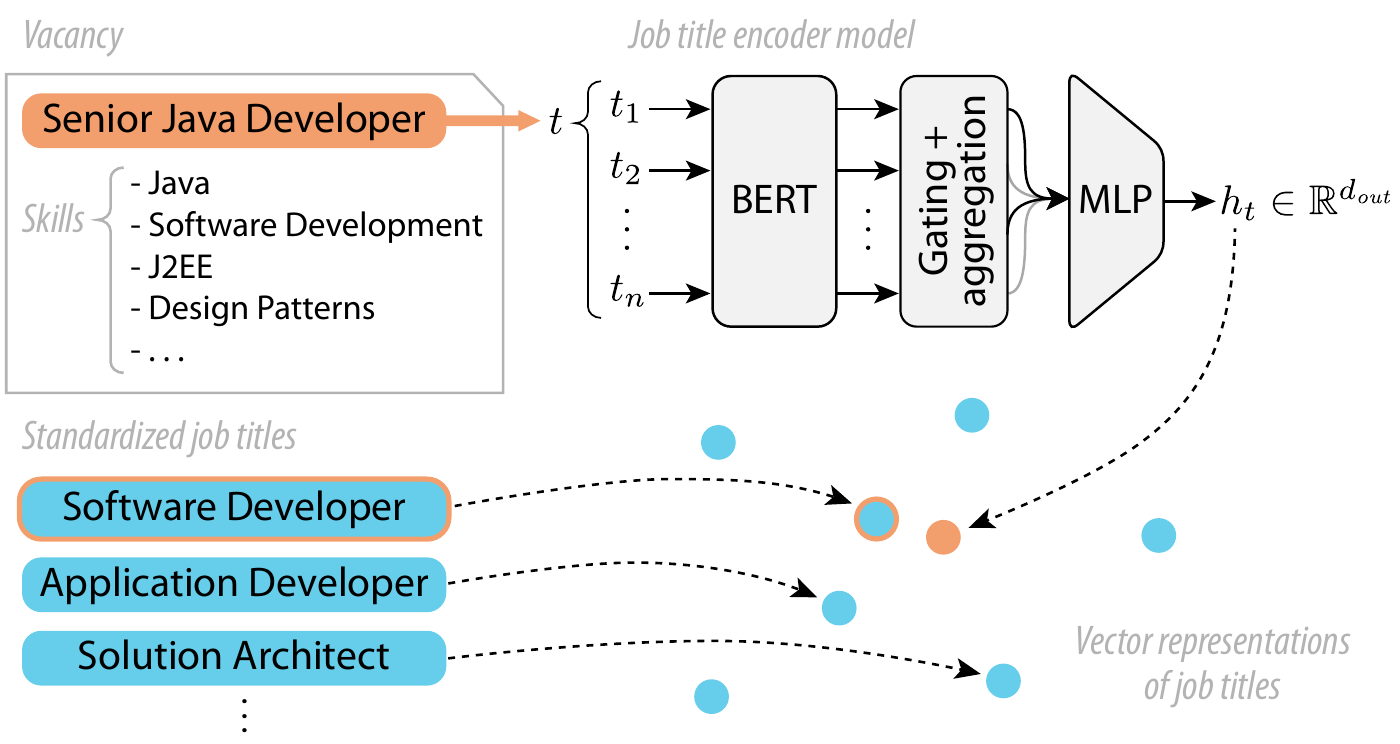}
\caption{Our JobBERT model uses a BERT-based encoder to obtain vector representations of job titles. Matching a job title to standardized job titles is based on proximity in the representation's vector space. Representations are trained based on vacancy titles and their associated skills: titles with similar skills are encoded in similar vectors.} \label{fig:overview}
\end{figure}

Job title normalization has traditionally been approached as a supervised learning problem.
A first effort by LinkedIn defined a ``couple dozen" standard job categories and performed classification based on common key phrases~\cite{bekkerman2011high}.
A more elaborate taxonomy of over 4k job titles was used in Carotene, using a hierarchical SVM-kNN cascade model for classification~\cite{javed2015carotene}.
The same task was more recently solved by DeepCarotene, using an end-to-end deep convolutional neural network~\cite{wang2019deepcarotene}.
This supervised approach suffers from three disadvantages:

\begin{enumerate}[(a)]
\item It requires an extensive set of job titles that are labeled with their normalized counterpart. The list of standardized job titles often has multiple thousands of items, making this annotation step extremely difficult without deeply familiar expert knowledge.\label{prob:a}
\item To keep the model up to date with trends in the job market, a continuous data labeling effort is required.\label{prob:b}
\item The resulting system is inherently tied to an a priori defined (and thus static) list of standardized titles. This limits its practical usability as it cannot be customized or tuned, for example towards specific HR department needs.\label{prob:c}
\end{enumerate}

The main contribution of this paper is a novel method for job title normalization that overcomes the above-mentioned problems \ref{prob:a}--\ref{prob:c}. 
We achieve this by approaching the task as a semantic text similarity (STS) task in which semantically similar job titles should lie close to each other in a learned vector space, as illustrated in \figref{fig:overview}. 
The resulting representations allow the usage of distance-based metrics for normalizing job titles with respect to \emph{any} job title taxonomy, thus alleviating problem~\ref{prob:c}.
This idea is similar to the approach taken in~\cite{neculoiu2016learning}, where a Siamese LSTM model is trained on explicit pairs of similar job titles. 
Although effective, their method still suffers from problems~\ref{prob:a} and~\ref{prob:b} as it requires a large handcrafted dataset of synonymous job titles, along with hand-picked augmentation rules. 
To bypass the need for labeling training data, we propose a distant supervision setup in which our semantic job title encoder learns to understand the meaning of job titles through the \emph{skills} they rely on, inferred from vacancies including descriptions thereof.
This approach and the reasoning behind it are further detailed in \secref{sec:method}.
Finally, we construct an evaluation dataset for the job title normalization task and report several performance metrics for our method as well as other commonly used sentence encoding techniques. 
We conclude that our JobBERT method outperforms Sentence-BERT (SBERT), which represents the state-of-the-art for general STS tasks~\cite{reimers-2019-sentence-bert}. 
To foster follow-up research on the task of job title normalization, we release the evaluation dataset to allow benchmarking.

\section{Job Title Representation Learning}
\label{sec:method}

Our method for constructing a representation $h_t$ for a job title $t$ is built upon the premise that skills are the essential components defining the job.
As such, semantically similar job titles should be similar in terms of the skills they require.
We apply this idea to our model by optimizing it to predict the skills that represent a given job title.
To bring this method into practice, a large corpus of titles and their corresponding set of skills is needed: we assume a vocabulary of skills $s \in V$, and a set of titles $t \in T$ each with their corresponding set of skills $S_t \subset V$.
Although this might seem like a hard requirement at first, this is easy in practice due to the generally high availability of \emph{vacancies} posted online. 
Vacancy titles are generally representative of the real distribution of job titles in the job market. Their description contains information about the skills that are required for the job. 
To avoid the costly manual tagging of skills within vacancies, we apply a simple distant supervision rule to create a noisy training set of vacancy titles with associated skills, through literal string matches in the vacancy texts with an extensive proprietary reference list of skills.

What follows is a description of our model architecture and optimization procedure. 
The main part of the model architecture is the job title encoder, as outlined in \figref{fig:overview}. 
It consists of the BERT model~\cite{devlin-etal-2019-bert}, a weighted averaging operation and a fully connected output network (MLP).
The final distributed representation $h_t \in \mathbb{R}^{d_\textit{out}}$ for job title $t$ is calculated as shown in \equref{eq:encoder}.
Newly introduced notations are explained below, along with the corresponding parts of the encoder network.

\begin{equation}\label{eq:encoder}
    h_t \;\;=\;\; \MLP\left(\sum_{i=1}^{n} \frac{\sigma\left(w_g \cdot \text{BERT}\left(t_i\right) + b_g\right)}{\sum_{j=1}^{n} \sigma\left(w_g \cdot \text{BERT}\left(t_j\right) + b_g\right)} \> \text{BERT}\left(t_i\right) \right)
\end{equation}

\subsubsection{Pre-trained Language Model}

We use BERT as the backbone of our encoder network. For efficiency reasons, we use the BERT-base model. It uses the WordPiece tokenizer to split an input text $t$ into a sequence of sub-word tokens $t_1, t_2, \ldots, t_n$, where $n$ depends on the length of the input~\cite{wu2016google}. This model outputs a $d$-dimensional vector for each input token $t_i$, denoted by $\text{BERT}(t_i)$.

\vspace{-0.3cm}

\subsubsection{Gating and Aggregation}

To obtain a fixed-length job title representation, the output vectors of BERT have to be aggregated. 
The authors of~\cite{reimers-2019-sentence-bert} show that averaging the token embeddings yields the best performance in their Sentence-BERT model, so we take this as our starting point.
However, we should note that not all words in a vacancy title provide information about the job content. 
As an example, the title \emph{``software developer London''} should lie close in the vector space to \emph{``software dev.\ in Ghent''}, but far from \emph{``business developer London''}. 
In other words, the model should learn to attribute more weight to meaningful words while neglecting indicators such as locations that are irrelevant with respect to job title classification. 
To this end, we add a gating mechanism that acts as an appropriate inductive bias to allow the model to ignore particular words before the aggregation into a single vector representation.
The gating coefficient for each token position $i$ is calculated as the weighted sum $(w_g \cdot \text{BERT}\left(t_i\right) + b_g)$ of all components in the token representation $\BERT(t_i)$, with the weight vector $w_g$ and bias $b_g$, followed by a sigmoid nonlinearity.
These gating coefficients are then normalized by their sum, after which they are used to aggregate all token representations into a single weighted sum. 

\vspace{-0.3cm}

\subsubsection{Projection MLP}

The last part of the encoder network is the projection multilayer perceptron (MLP), which consists of two linear layers with a ReLU activation in between. It transforms the aggregated $d$-dimensional embedding into the final representation $h_t \in \mathbb{R}^{d_\textit{out}}$.

\vspace{-0.3cm}

\subsubsection{Context Matrix and Optimization}

The encoder network is trained to predict the bag of skills $S_t$ that corresponds to a given title $t$ in the training data. 
To formalize this, we draw inspiration from the Skip-Gram model with negative subsampling~\cite{mikolov2013distributed}. 
For each positive training pair $(t, s_j)$, with $s_j \in S_t$ a skill associated with job title $t$, we randomly draw $K$ negative skills from a noise distribution.
This distribution is defined by the frequency distribution of the skills in the training corpus raised to the power of $3 / 4$, as this has been shown to work well~\cite{mikolov2013distributed}.
The model is optimized to predict the presence of the positive label $s_j$ while predicting the absence of the $K$ negative labels. 
To facilitate this optimization, the model contains a trainable context matrix $U\in\mathbb{R}^{|V| \times d_\textit{out}}$ that consists of context vectors $u_i \in \mathbb{R}^{d_\textit{out}}$ for each skill $s_i$ in the skill vocabulary $V$.\footnote{Note that the context matrix is not shown in \figref{fig:overview} as it is only needed during training and can be discarded at inference time.}
The log-likelihood of each positive pair $(t, s_j)$ (with $s_j \in S_t)$ is given by:
\begin{equation}
    \mathcal{J}(t, s_j) = \log \sigma\left(u_{j} \cdot h_t \right)+\sum_{k=1}^{K} \log \sigma\left(-\tilde{u}_{k} \cdot h_t\right)
\end{equation}

In the above equation, $\left\{\tilde{u}_{k} \mid k=1 \ldots K\right\}$ are the context vectors corresponding to the $K$ negative sampled skills.
The gradient updates for both the encoder model and the context matrix are calculated with the objective of minimizing the negative log-likelihood.

\section{Experimental Setup}
\label{sec:experiments}

\subsubsection{Benchmark Dataset}

In order to evaluate the performance of our model, we create and publish a new dataset of vacancy titles that are labeled with the standardized ESCO occupations.\footnote{\url{https://github.com/jensjorisdecorte/JobBERT-evaluation-dataset}} 
This data was gathered from a large governmental online job board, on which each job posting is tagged with the most suitable ESCO occupation label by its creator.\footnote{\url{https://www.myfuturejobs.gov.my/}} 
The ESCO occupations are organized within a hierarchical structure. 
For example, both \textit{``aircraft engine assembler''} and \textit{``wave soldering machine operator''} are leaf nodes in the sub-tree of the generic \textit{``assembler''} occupation.
To have a mutually exclusive label space, we reduce the hierarchy of ESCO occupations into a list of its leaf nodes, after which 2,675 out of the total of 2,942 occupations remain. 
We discard all non-English vacancy titles in the dataset by applying the fastText Large language identification model~\cite{joulin2016fasttext,joulin-etal-2017-bag}. 
The final dataset contains a total of 30,926 unique English vacancy titles, each tagged with one of the remaining ESCO occupations.
We make a stratified split of this dataset based on the ESCO label distribution into two equally sized parts, for validation and testing.
We note that not all vacancy titles in this dataset are equally descriptive.
Hence, from analyzing a random subsample of 2,000 vacancies in this dataset, we estimate an upper bound of 65\% of the vacancy titles that can be linked to their corresponding ESCO label without ambiguity.

\vspace{-0.3cm}

\subsubsection{Training Details}

We describe the high-level experimental setup here and refer to \appref{app:details} for more details. 
We select a large training corpus of 300 million English vacancies from TechWolf's internal Data Lake.
A proprietary vocabulary of over 35k clean skill names is used to generate the skill tags for self-supervision.
Vacancies are tagged with skills based on literal occurrence in the text.\footnote{Casing is ignored, except for skills that contain a majority of uppercase characters.}
This is a rather simple skill extraction method that can be improved upon in many ways.
However, as a distant supervision approach applied on a large dataset, it is qualitative enough to provide strong results.

The BERT model is initialized with its pre-trained weights. 
We choose an output dimension of $d_{out}$\,=\,300, which is much smaller than the original 768-dimensional BERT embeddings.
This is beneficial for the application of job title normalization as it reduces storage requirements and speeds up the calculation of similarity metrics.
We use a batch size of 64 (title, skill)-pairs during training. 
The number of negatives appeared to have little impact on the performance and we set it to $K$\,=\,5.
In our first experiment, the pre-trained BERT weights are kept fixed. 
Afterwards, we investigate the further improvements to be made by also finetuning the BERT weights.
Throughout the training, the performance is measured on the validation set in terms of mean reciprocal rank (MRR) after each multiple of 5k batches.
We stop training when the validation score shows no further improvements, which happens after the first 8.5\% of the fist epoch.

\section{Results and Analysis}
\label{sec:results}

We compare our JobBERT model with most commonly used sentence representation models LASER~\cite{DBLP:journals/tacl/ArtetxeS19}, fastText~\cite{joulin-etal-2017-bag} and Sentence-BERT~\cite{reimers-2019-sentence-bert}.
For the Sentence-BERT model, we used its base model with averaged token embeddings (SBERT\textsubscript{\textsc{avg}}).
We also included the plain BERT-base model~\cite{devlin-etal-2019-bert} with averaged token embeddings (BERT\textsubscript{\textsc{avg}}).
We calculate both MRR and Recall@k metrics on the test set by ranking the ESCO labels based on cosine similarity with the input title.
In \tabref{tab:results} we report both macro and micro averages because of the long tail class distribution in the test dataset.
Our results confirm the findings of~\cite{reimers-2019-sentence-bert} that averaging BERT token embeddings does not yield good representations, as they are surpassed by both fastText and LASER representations.
However, our JobBERT approach outperforms Sentence-BERT, even when keeping the pre-trained BERT weights fixed, while finetuning (\textsc{ft})  BERT  further considerably increases the test scores.

We finally note that our JobBERT model can be inspected by analyzing the gating coefficients for each token in a given job title. Such analysis in \appref{app:weights} demonstrates that the model successfully neglects words that are irrelevant for job title classification (\eg locations).

\vspace{-0.6cm}

\begin{table*}[bht]
\fontsize{9pt}{9pt}\selectfont
    \centering
    \begin{threeparttable}
    \begin{tabular}{l cc cc cc cc}
    \toprule
    & \multicolumn{2}{c}{\textbf{MRR}} & \multicolumn{2}{c}{\textbf{Recall@1}} & \multicolumn{2}{c}{\textbf{Recall@5}} & \multicolumn{2}{c}{\textbf{Recall@10}} \\
    \cmidrule(lr){2-3} \cmidrule(lr){4-5} \cmidrule(lr){6-7} \cmidrule(lr){8-9}
    \textbf{Method} & Macro & Micro & Macro & Micro & Macro & Micro & Macro & Micro \\
    \midrule
    LASER~\cite{DBLP:journals/tacl/ArtetxeS19} & 0.2194 & 0.2250 & 0.1600 & 0.1723 & 0.2596 & 0.2537 & 0.3046 & 0.2960\\
    fastText~\cite{joulin-etal-2017-bag} & 0.2533 & 0.2313 & 0.1710 & 0.1602 & 0.3217 & 0.2823 & 0.3918 & 0.3443\\
    BERT\textsubscript{\textsc{avg}} & 0.2000 & 0.2056 & 0.1355 & 0.1486 & 0.2426 & 0.2396 & 0.2963 & 0.2843\\
    SBERT\textsubscript{\textsc{avg}} & 0.2695 & 0.2648 & 0.1863 & 0.1930 & 0.3426 & 0.3262 & 0.4160 & 0.3801\\
    \midrule
    JobBERT\tnote{*} & 0.3261 & 0.2690 & 0.2307 & 0.1919 & 0.4207 & 0.3352 & 0.4991 & 0.3993 \\
    JobBERT\textsubscript{\textsc{ft}}\tnote{*}\phantom{x}  & \textbf{0.3641} & \textbf{0.3092} & \textbf{0.2666} & \textbf{0.2248} & \textbf{0.4632} & \textbf{0.3865} & \textbf{0.5440} & \textbf{0.4604} \\
    \bottomrule
    \end{tabular}
    \begin{tablenotes}
    \small
    \item[*]: Our proposed methods; \textsc{ft} = with finetuning the BERT weights.\\
    \end{tablenotes}
    
    \end{threeparttable}
    \caption{Comparison of representation-based job title classification methods. 
    }
    \label{tab:results}
\end{table*}

\vspace{-0.9cm}

\section{Conclusion}
\label{sec:conclusion}

We present JobBERT, a novel approach for the task of job title normalization, which builds upon the premise that skills are the essential components defining a job. We rely on distant supervision by learning to represent vacancy titles based on skills included in their description.
This eliminates the need for manual data labeling, which is costly, especially to keep the model up to date with rapidly changing trends in the job market. 
The strength of the JobBERT representations is demonstrated by applying them for the task of job title normalization through mapping on a taxonomy not seen during training of the model.
The task is cast as a ranking problem, and we achieve improvements of the order of +10\% in mean reciprocal rank as well as Recall@10 compared to state-of-the-art encoders for semantic text similarity.

\section{Acknowledgments}

We thank the anonymous reviewers for their valuable feedback. This project was funded by the Flemish Government, through Flanders Innovation \& Entrepreneurship (VLAIO, project HBC.2020.2893).

%
%
%
%

\bibliographystyle{splncs04}
\bibliography{references}

\begin{subappendices}
\renewcommand{\thesection}{\Alph{section}}%

\section{Training details}\label{app:details}

During training, we keep all (title, skill)-pairs originating from the same vacancy together in one batch and pass each unique title only once through the encoder model. Given that an average vacancy in our training data contains around 14.7 literal skill mentions, this trick speeds up the forward pass during training by over an order of magnitude.
We use a relatively large batch size of 64 which makes the training procedure robust to noise introduced by the distant supervision.

In the first experiment, we initialize BERT with its pre-trained weights and keep those fixed.
The context matrix $U$ is initialized with zero values, in accordance with the original implementation of the Skip-Gram model~\cite{mikolov2013distributed}.
The token gating and the MLP both consist of linear layers.
Each of those linear layers --- with input dimension $I$ and output dimension $O$ --- are initialized as follows:
\begin{enumerate}
    \item The \emph{weight} parameters are randomly initialized from a uniform distribution in the range of $\left(-\sqrt{\frac{1}{I}}, \sqrt{\frac{1}{I}}\right)$,
    \item The \emph{bias} parameters are randomly initialized from a uniform distribution in the range of $\left(-\sqrt{\frac{1}{O}}, \sqrt{\frac{1}{O}}\right)$.
\end{enumerate}
We use a simple stochastic gradient descent (SGD) optimizer with a learning rate of 0.05 for both the gating mechanism and MLP weights.
The context matrix is optimized with the Adaptive optimization scheme Adagrad~\cite{JMLR:v12:duchi11a} with the default learning rate of 0.01.
This optimizer tracks which skills occur more frequently during training and reduces the learning rate for those, which stabilizes the training procedure.

In the second experiment, the BERT weights are finetuned.
In order to stabilize the training procedure, we initialize the model with the weights from the first experiment.
The learning rate of the SGD optimizer is reduced to 0.001 as it is now also used to update the BERT weights.

For the fixed BERT model, we see no significant performance improvement on the evaluation set after training on 15 million vacancies.
Further improvements from finetuning the BERT weights stall after training on an additional 10.5 million vacancies.
As such, we train for less than one epoch in total, on only the most recent 8.5\% of vacancies.
All experiments were run on an NVIDIA K80 GPU.

\section{Analysis of the Gating Coefficients}\label{app:weights}

The token gating mechanism in our JobBERT model allows the inspection of which tokens in a job title the model perceives to be important for its meaning. Doing so reveals that the model successfully ignores tokens that do not communicate information regarding the job content. Below, some examples are illustrated. Each token in the input job title is accompanied by its gating coefficient. The coefficient values range from zero (not important) to one (important) and are reflected in the intensity of the red background color:

\begin{enumerate}\small\setstretch{2.0}

\item \colorbox{red!76}{elementary\textsubscript{(.77)}}\colorbox{red!83}{school\textsubscript{(.84)}}\colorbox{red!94}{teacher\textsubscript{(.95)}}\colorbox{red!35}{french\textsubscript{(.35)}}\colorbox{red!1}{part\textsubscript{(.02)}}\colorbox{red!1}{-\textsubscript{(.01)}}\colorbox{red!7}{time\textsubscript{(.07)}}.\label{ex:1}

\item \colorbox{red!97}{AI\textsubscript{(.97)}}\colorbox{red!50}{specialist\textsubscript{(.51)}}\colorbox{red!7}{in\textsubscript{(.07)}}\colorbox{red!18}{London\textsubscript{(.18)}}.\label{ex:2}

\item \colorbox{red!43}{head\textsubscript{(.43)}}\colorbox{red!26}{of\textsubscript{(.26)}}\colorbox{red!89}{data\textsubscript{(.90)}}\colorbox{red!80}{science\textsubscript{(.80)}}\colorbox{red!9}{in\textsubscript{(.09)}}\colorbox{red!12}{our\textsubscript{(.13)}}\colorbox{red!11}{head\textsubscript{(.11)}}\colorbox{red!9}{quarters\textsubscript{(.09)}}.\label{ex:3}

\end{enumerate}

Examples~\ref{ex:1} and~\ref{ex:2} illustrate how meaningless indicators such as ``in London'' or ``part-time'' are ignored by the model, while the important tokens receive high gating coefficients. Example~\ref{ex:3} contains the ``head'' token twice, but attributes a four times higher gating coefficient to the first appearance which adds to the meaning of the job title.

\end{subappendices}

\end{document}